\documentclass[sigconf]{acmart}

\AtBeginDocument{%
  \providecommand\BibTeX{{%
    \normalfont B\kern-0.5em{\scshape i\kern-0.25em b}\kern-0.8em\TeX}}}

\copyrightyear{2022} 
\acmYear{2022} 
\setcopyright{acmcopyright}\acmConference[WWW '22 Companion]{Companion Proceedings of the Web Conference 2022}{April 25--29, 2022}{Virtual Event, Lyon, France}
\acmBooktitle{Companion Proceedings of the Web Conference 2022 (WWW '22 Companion), April 25--29, 2022, Virtual Event, Lyon, France}
\acmPrice{15.00}
\acmDOI{10.1145/3487553.3524258}
\acmISBN{978-1-4503-9130-6/22/04}

\usepackage{subcaption}
\usepackage{caption}
\usepackage{multirow}
\usepackage{enumitem}
\usepackage{arydshln}

\begin{document}

\title{Universal Graph Transformer Self-Attention Networks}

\author{Dai Quoc Nguyen}
\authornote{This work was done before Dai Quoc Nguyen joined Oracle Labs, Australia. UGformer Variant 1 was proposed when Dai Quoc Nguyen was a PhD student at Monash University, Australia.}
\affiliation{\institution{Oracle Labs}\country{Australia}}
\email{dai.nguyen@oracle.com}

\author{Tu Dinh Nguyen}
\affiliation{\institution{VinAI Research}\country{Vietnam}}
\email{v.tund21@vinai.io}

\author{Dinh Phung}
\affiliation{\institution{Monash University}\country{Australia}}
\email{dinh.phung@monash.edu}



\begin{abstract}

We introduce a transformer-based GNN model, named UGformer, to learn graph representations. In particular, we present two UGformer variants, wherein the first variant (publicized in  September 2019) is to leverage the transformer on a set of sampled neighbors for each input node, while the second (publicized in May 2021) is to leverage the transformer on all input nodes. Experimental results demonstrate that the first UGformer variant achieves state-of-the-art accuracies on benchmark datasets for graph classification in both inductive setting and unsupervised transductive setting; and the second UGformer variant obtains state-of-the-art accuracies for inductive text classification. The code is available at: \url{https://github.com/daiquocnguyen/Graph-Transformer}.

\end{abstract}

\maketitle

\section{Introduction}

A graph is a connected network of nodes and edges. 
This type of graph-structured data is a fundamental mathematical representation and ubiquitous. 
They found applications in virtually all aspects of our daily lives from pandemic outburst response, internet of things, drug discovery to circuit design, to name a few.
In machine learning and data science, learning and inference from graphs have been one of the most trending research topics. 
However, as data grow unprecedentedly in volume and complexity in modern time, traditional learning methods for graph are mostly inadequate to model increasing complexity, to harness rich contextual information as well as to scale with large-scale graphs. 
The recent rise of deep learning, and in turn, of representation learning field has radically advanced machine learning research in general, and pushing the frontier of graph learning. 
In particular, the notion of graph representation learning has recently emerged as a new promising learning paradigm, which aims to learn a parametric mapping function that embeds nodes, subgraphs, or the entire graph into low-dimensional continuous vector spaces \citep{hamilton2017representation,wu2019comprehensive,NGUYEN2021Thesis}.
The central challenge to this endeavor is to learn rich classes of complex functions to capture and preserve the graph structural information as much as possible and also be able to geometrically represent the structural information in the embedded space.

Recently, graph neural networks (GNNs) become an essential strand, forming the third direction to learn low-dimensional continuous representations for nodes and graphs \citep{scarselli2009graph,hamilton2017representation,wu2019comprehensive}.
In general, GNNs use an aggregation function to update the vector representation of each node by transforming and aggregating the vector representations of its neighbours \citep{kipf2017semi,hamilton2017inductive,velickovic2018graph,Nguyen2020QGNN}.
Then GNNs apply a graph-level readout function such as simple sum pooling to obtain graph embeddings \citep{Gilmer2017,zhang2018end,Ying2018diffpool,verma2018graph,xu2019powerful}.
GNN-based approaches provide faster and practical training and state-of-the-art results on benchmark datasets for downstream tasks such as node classification \citep{kipf2017semi}, graph classification \citep{xu2019powerful}, knowledge graph completion \citep{Nguyen2022NoGE}, vulnerability detection \citep{Nguyen2021regvd}, and text classification \citep{yao2019graph}.
To further improve the performance, it is worth developing advanced GNNs to better update vector representations of nodes from their neighbours.
Currently, there are novel applications of the transformer \citep{vaswani2017attention} recognized, published, and used successfully in natural language processing.
Inspired by this fact, we consider the use of the transformer to a new domain such as GNNs as a novelty and present UGformer, a transformer-based GNN model, to learn graph representations. 
Our main contributions in this paper are as follows:

$\bullet$ We propose a transformer-based GNN model, named UGformer, to learn graph representations. 
In particular, we consider two model variants of (i) leveraging the transformer on a set of sampled neighbors for each input node and (ii) leveraging the transformer on all input nodes.
\textit{We publicized the first UGformer variant in September 2019 and the second UGformer variant in May 2021.}

$\bullet$ The unsupervised learning is essential in both industry and academic applications, where expanding unsupervised GNN models is more suitable to address the limited availability of class labels. Thus, we present an \textit{unsupervised transductive} learning approach to train GNNs.

$\bullet$ Experimental results show that the first UGformer variant obtains state-of-the-art accuracies on social network and bioinformatics datasets for graph classification in both inductive setting and unsupervised transductive setting; and the second UGformer variant produces state-of-the-art accuracies on benchmark datasets for inductive text classification.

\section{The proposed UGformer}
\label{sec:ourmodel}

\subsection{Variant 1: Leveraging the transformer on a set of sampled neighbors for each node}

\begin{figure}[!ht]
\centering
\includegraphics[width=0.45\textwidth]{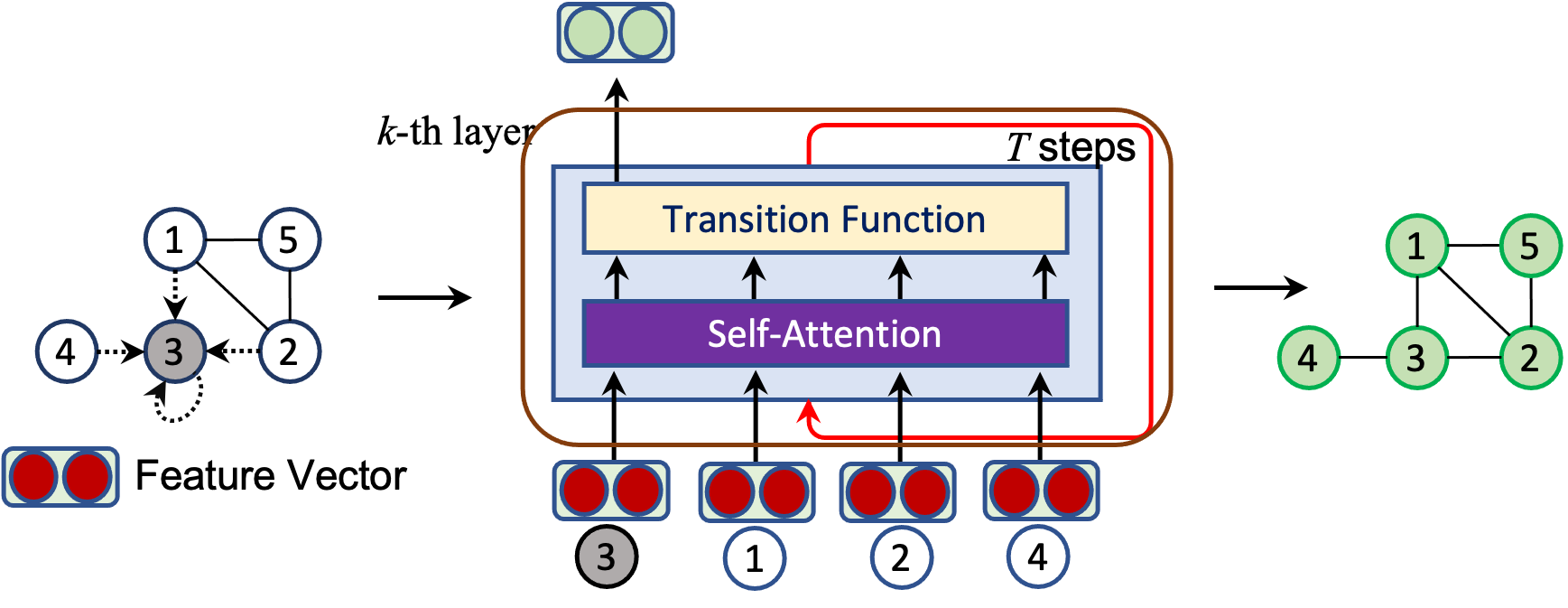}
\captionof{figure}{An illustration of UGformer Variant 1.}
\label{fig:UGformer}
\end{figure}

We publicized the first UGformer variant in September 2019. 
This variant is also suitable for large graphs.
Given an input graph $\mathcal{G}$, we uniformly sample a set $\mathcal{N}_\mathsf{v}$ of neighbors for each $\mathsf{v} \in \mathcal{V}$ and then input $\mathcal{N}_\mathsf{v}\cup\left\{\mathsf{v}\right\}$ to the UGformer learning process, as illustrated in Figure \ref{fig:UGformer}. 
Note that we sample a different $\mathcal{N}_\mathsf{v}$ for node $\mathsf{v}$ at each training batch.
We utilize the transformer to construct multiple layers stacked on top of each other, wherein the weight matrices are shared in both the self-attention and transition functions across all positions and timesteps.
In particular, regarding the $k$-th layer, given a node $\mathsf{v} \in \mathcal{V}$, at each step $t$, we induce a transformer-based function to aggregate the vector representations for all nodes $\mathsf{u} \in \mathcal{N}_\mathsf{v}\cup\left\{\mathsf{v}\right\}$ as follows:
\begin{eqnarray}
\boldsymbol{\mathsf{x}}^{(k)}_{t,\mathsf{u}} &=& \textsc{LayerNormalization}\left(\boldsymbol{\mathsf{h}}^{(k)}_{t-1,\mathsf{u}} + \textsc{ATT}\left(\boldsymbol{\mathsf{h}}^{(k)}_{t-1,\mathsf{u}}\right)\right) \label{equa:usa2} \\
\boldsymbol{\mathsf{h}}^{(k)}_{t,\mathsf{u}} &=& \textsc{LayerNormalization}\left(\boldsymbol{\mathsf{x}}^{(k)}_{t,\mathsf{u}} + \textsc{Trans}\left(\boldsymbol{\mathsf{x}}^{(k)}_{t,\mathsf{u}}\right)\right) \label{equa:usa1}
\end{eqnarray}
where $\boldsymbol{\mathsf{h}}^{(0)}_{0,\mathsf{v}} = \boldsymbol{\mathsf{h}}_{\mathsf{v}}^{(0)}
$ is the feature vector of $\mathsf{v}$;
$\textsc{Trans}(.)$ and $\textsc{ATT}(.)$ denote a MLP network and a self-attention layer respectively.
In particular, we have:
\begin{equation} 
\textsc{ATT}\left(\boldsymbol{\mathsf{h}}^{(k)}_{t-1,\mathsf{u}}\right) = \sum_{\mathsf{u}' \in \mathcal{N}_\mathsf{v}\cup\left\{\mathsf{v}\right\}}\alpha^{(k)}_{\mathsf{u},\mathsf{u}'}\left(\boldsymbol{V}^{(k)}\boldsymbol{\mathsf{h}}^{(k)}_{t-1,\mathsf{u}'}\right) 
\label{equa:usatt1}
\end{equation}
where $\boldsymbol{V}^{(k)} \in \mathbb{R}^{d\times d}$ is a value-projection weight matrix; $\alpha_{\mathsf{u},\mathsf{u}'}$ is an attention weight, which is computed using the $\mathsf{softmax}$ function over scaled dot products between nodes $\mathsf{u}$ and $\mathsf{u}'$:
\begin{equation}
\alpha^{(k)}_{\mathsf{u},\mathsf{u}'} = \mathsf{softmax}\left(\frac{\left(\boldsymbol{Q}^{(k)}\boldsymbol{\mathsf{h}}^{(k)}_{t-1,\mathsf{u}}\right)^\mathsf{T}\left(\boldsymbol{K}^{(k)}\boldsymbol{\mathsf{h}}^{(k)}_{t-1,\mathsf{u}'}\right)}{\sqrt{d}}\right) 
\end{equation}
where $\boldsymbol{Q}^{(k)} \in \mathbb{R}^{d\times d}$ and $\boldsymbol{K}^{(k)} \in \mathbb{R}^{d\times d}$ are query-projection and key-projection matrices, respectively.
\begin{equation}
\boldsymbol{\mathsf{H}}_t^{(k)} = \textsc{Attention}_{{\mathcal{N}_\mathsf{v}\cup\{\mathsf{v}\}}}\left(\boldsymbol{\mathsf{H}}_{t-1}^{(k)}\boldsymbol{Q}^{(k)}, \boldsymbol{\mathsf{H}}_{t-1}^{(k)}\boldsymbol{K}^{(k)}, \boldsymbol{\mathsf{H}}_{t-1}^{(k)}\boldsymbol{V}^{(k)}\right)
\end{equation}

After $T$ steps, we feed $\boldsymbol{\mathsf{h}}^{(k)}_{T,\mathsf{v}} \in \mathbb{R}^d$ to the next $(k+1)$-th layer as:
\begin{equation}
\boldsymbol{\mathsf{h}}_{\mathsf{v}}^{(k+1)} = \boldsymbol{\mathsf{h}}^{(k+1)}_{0,\mathsf{v}} = \boldsymbol{\mathsf{h}}^{(k)}_{T,\mathsf{v}} , \forall \mathsf{v} \in \mathcal{V}
\label{equa:nextlayerequa}
\end{equation}
If we do not share the weight matrices in both the self-attention and transition functions across all positions and timesteps, $T$ becomes the number of self-attention layers within each UGformer layer.

\subsection{Variant 2: Leveraging the transformer on all input nodes}

\begin{figure}[!ht]
\centering
\includegraphics[width=0.35\textwidth]{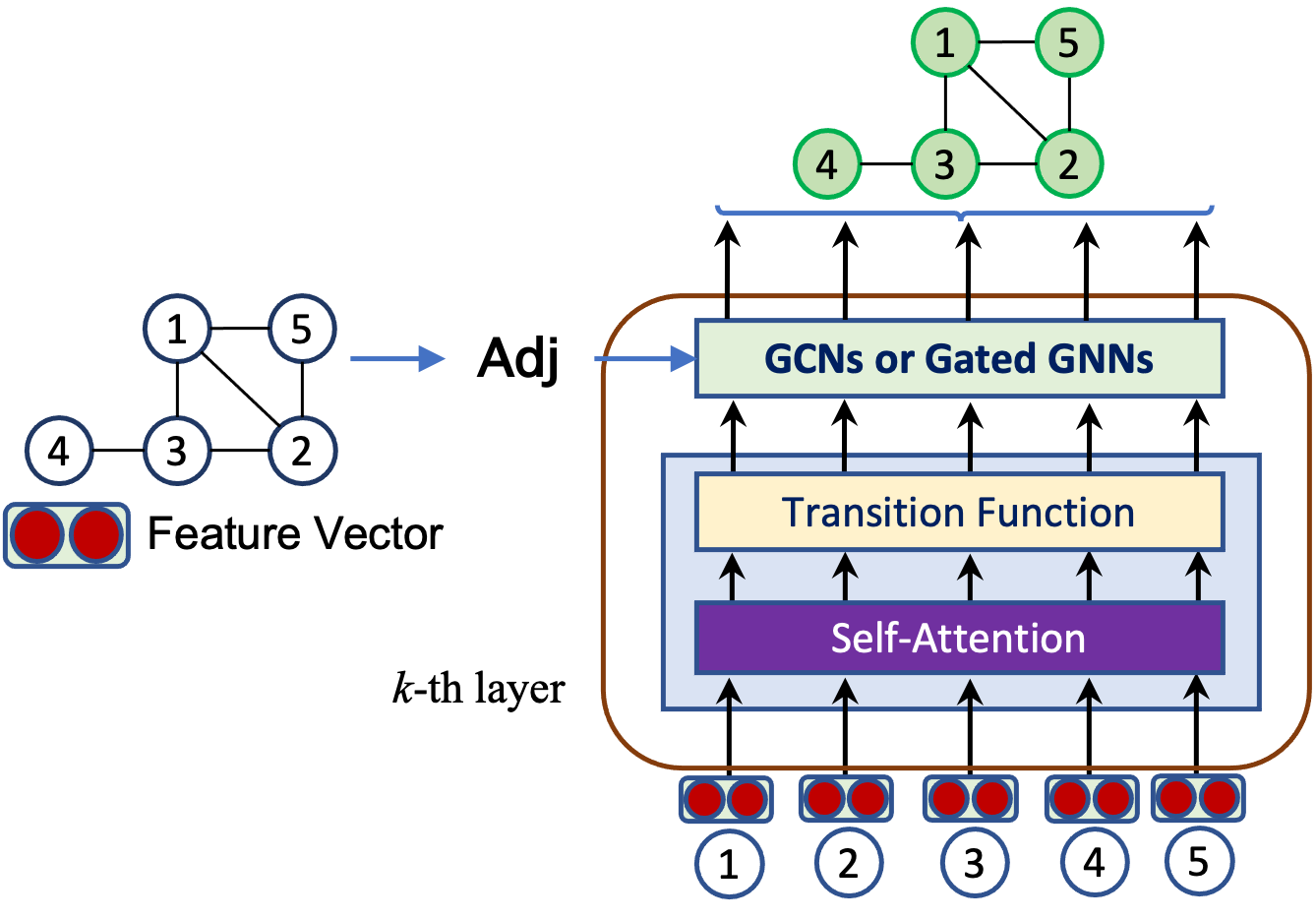}
\captionof{figure}{An illustration of UGformer Variant 2.}
\label{fig:UGformerv2}
\end{figure}

We publicized the second UGformer variant in May 2021, when we aimed to leverage the transformer for small and medium graphs.
It is worth noting that all links/interactions among all positions in the self-attention layer build up a {complete} network. 
Hence, if we apply \textit{only} the self-attention mechanism to all nodes of the input graph $\mathcal{G}$, we ignore the structure of $\mathcal{G}$.
To overcome this limitation, we propose that each UGformer layer consists of a transformer self-attention network followed by a GNN layer such as GCNs \citep{kipf2017semi} or Gated GNNs \citep{li2015gated}. This model variant 
combines the transformer with the graph structure to learn better graph representations.
Formally, we define a UGformer layer as illustrated in Figure \ref{fig:UGformerv2} as:
\begin{eqnarray}
\boldsymbol{\mathsf{H'}}^{(k)} &=& \textsc{Attention}_{\mathcal{V}}\left(\boldsymbol{\mathsf{H}}^{(k)}\boldsymbol{Q}^{(k)}, \boldsymbol{\mathsf{H}}^{(k)}\boldsymbol{K}^{(k)}, \boldsymbol{\mathsf{H}}^{(k)}\boldsymbol{V}^{(k)}\right) \label{equa:onlystransformer}\\
\boldsymbol{\mathsf{H}}^{(k+1)} &=& \mathsf{GNN}\left(\textbf{A}, \boldsymbol{\mathsf{H'}}^{(k)}\right) \label{equa:gnnlayer}
\end{eqnarray}
where $\boldsymbol{\mathsf{H}}^{(0)}$ is the feature matrix of all nodes in $\mathcal{G}$, and $\textbf{A}$ is the adjacency matrix.
Equation \ref{equa:onlystransformer} is with respect to Equations \ref{equa:usa2} and \ref{equa:usa1} when $T=1$.

\section{Demonstration of UGformer}

\subsection{UGformer Variant 2 for inductive text classification}

We demonstrate the advantage of UGformer for inductive text classification.
Firstly, we follow \citep{zhang2020every,Nguyen2020QGNN} to build a graph $\mathcal{G}$ for each textual document by representing unique words as nodes and co-occurrences between words (within a fixed-size sliding window of length 3) as edges. 
We then employ our Variant 2, wherein we adapt a Gated GNN layer \citep{li2015gated} for Equation \ref{equa:gnnlayer} as the Gated GNN layer is more suitable for our data.
After that, to produce the graph embedding $\boldsymbol{\mathsf{e}}_{\mathcal{G}}$, we follow \citep{Nguyen2021regvd} to construct a graph-level readout function as:
\begin{eqnarray}
\boldsymbol{\mathsf{e}}_\mathsf{v} &=& \sigma\left(\textbf{w}^\mathsf{T}\boldsymbol{\mathsf{h}}_\mathsf{v}^{(K)}+\mathsf{b}\right)\odot\mathsf{g}\left(\textbf{W}_1\boldsymbol{\mathsf{h}}_{\mathsf{v}}^{(K)}+\boldsymbol{\mathsf{b}}_1\right) \\
\boldsymbol{\mathsf{e}}_{\mathcal{G}} &=& \sum_{\mathsf{v} \in \mathcal{V}}\boldsymbol{\mathsf{e}}_\mathsf{v}\odot\mathsf{max\_pooling}\left\{\boldsymbol{\mathsf{e}}_\mathsf{v}\right\}_{\mathsf{v} \in \mathcal{V}}
\end{eqnarray}
where $\boldsymbol{\mathsf{h}}_\mathsf{v}^{(K)}$ is the vector representation of node $\mathsf{v}$ at the last $K$-th layer; and $\sigma\left(\textbf{w}^\mathsf{T}\boldsymbol{\mathsf{h}}_\mathsf{v}^{(K)}\right)$ acts as soft attention mechanisms over nodes.
Finally, we also feed $\boldsymbol{\mathsf{e}}_{\mathcal{G}}$ to a single fully-connected layer followed by a $\mathsf{softmax}$ layer to perform the text classification task.

\subsubsection{Experimental setup}

We follow \citep{yao2019graph,zhang2020every,Nguyen2020QGNN} to use four benchmarks -- MR, R8, R52, and Ohsumed. 
We follow \citep{zhang2020every,Nguyen2020QGNN} to use random vectors or pre-trained Glove \citep{pennington2014glove} with the dimension size of 300 to initialize the feature vectors. 
We also follow \citep{yao2019graph,zhang2020every,Nguyen2020QGNN} to build a 2-layer model. 
We set the number of attention heads to 2 and the hidden size to 384. We vary the learning rate in $ \{1e^{-4}, 5e^{-4}, 1e^{-3}, 5e^{-3}\}$. 
Then, we utilize the Adam optimizer to train the model up to 150 epochs to evaluate our trained model.
We also run 10 times and report the mean accuracy and standard deviation, wherein for each time, we randomly split 10\% text from the training set to have the validation set for hyper-parameter turning.

\begin{table}[!ht]
\centering
\caption{Text classification accuracies (\%) on the test sets w.r.t UGformer Variant 2. Some baseline results are taken from \citep{yao2019graph}.
Note that the original results of TextING \citep{zhang2020every} are reported for the best model on the test set; thus, we consider the results of TextING taken from \citep{Nguyen2020QGNN} for a fair comparison. 
``w/o GNN'' means that we do not use the GNN layer within the UGformer layer.
}
\resizebox{8.5cm}{!}{
\setlength{\tabcolsep}{0.85em}
\begin{tabular}{lcccc}
\hline 
{\bf Dataset} & \textbf{MR} & \textbf{R8} & \bf R52 & \bf Ohsumed \\
\hline
Bi-LSTM & 77.68 $\pm$ 0.86 & 96.31 $\pm$ 0.33 & 90.54 $\pm$ 0.91 & 49.27 $\pm$ 1.07 \\
fastText & 75.14 $\pm$ 0.20 & 96.13 $\pm$ 0.21 & 92.81 $\pm$ 0.09 & 57.70 $\pm$ 0.49 \\
TextGCN & 76.74 $\pm$ 0.20 & \textbf{97.07 $\pm$ 0.10} & 93.56 $\pm$ 0.18 & 68.36 $\pm$ 0.56 \\
TextING & 78.86 $\pm$ 0.26 & 96.90 $\pm$ 0.23 & 93.34 $\pm$ 0.24 & 69.72 $\pm$ 0.30 \\ 
TextQGNN & 78.93 $\pm$ 0.29 & 97.02 $\pm$ 0.28 & 94.45 $\pm$ 0.35 & 69.93 $\pm$ 0.31\\
\hline
\textbf{UGformer} & \textbf{79.29 $\pm$ 0.46} & 97.05 $\pm$ 0.32 & \textbf{94.71 $\pm$ 0.42} & \textbf{70.63 $\pm$ 0.18} \\
\hdashline
\ \ \ \ \ \ w/o GNN & 77.47 $\pm$ 0.51 & 96.50 $\pm$ 0.42 & 93.85 $\pm$ 0.08 & 68.27 $\pm$ 0.52\\
\hline
\end{tabular}
}
\label{tab:textclsresults}
\end{table}

\subsubsection{Main results}
Table \ref{tab:textclsresults} presents the classification accuracy results of our UGformer and the baseline models. In general, our UGformer outperforms the baseline models, produces state-of-the-art accuracies on three benchmark datasets R52, \textsc{Ohsumed}, and MR, and obtains a highly competitive accuracy on R8.
In Table \ref{tab:textclsresults}, we also compute and report our ablation results for using only the transformer w.r.t Equation \ref{equa:onlystransformer}, i.e., \textit{not} using the GNN layer within the UGformer layer w.r.t Equation \ref{equa:gnnlayer}; hence we have:
\begin{equation}
\boldsymbol{\mathsf{H}}^{(k+1)} = \textsc{Attention}_{\mathcal{V}}\left(\boldsymbol{\mathsf{H}}^{(k)}\boldsymbol{Q}^{(k)}, \boldsymbol{\mathsf{H}}^{(k)}\boldsymbol{K}^{(k)}, \boldsymbol{\mathsf{H}}^{(k)}\boldsymbol{V}^{(k)}\right) \nonumber
\end{equation}
It is worth noting that the scores degrade, thus showing the effectiveness of our proposed UGformer in integrating the transformer with the graph structure to learn better graph representations.

\subsection{UGformer Variant 1 for graph classification in an inductive setting}

Following \citep{xu2018representation,xu2019powerful}, we apply the vector concatenation across the layers to obtain the vector representations $\boldsymbol{\mathsf{e}}_\mathsf{v}$ of nodes $\mathsf{v}$ as:
\begin{equation}
\boldsymbol{\mathsf{e}}_\mathsf{v} = \left[\boldsymbol{\mathsf{h}}^{(1)}_{\mathsf{v}}\parallel \boldsymbol{\mathsf{h}}^{(2)}_{\mathsf{v}}\parallel ...\parallel \boldsymbol{\mathsf{h}}^{(K)}_{\mathsf{v}}\right] , \forall \mathsf{v} \in \mathcal{V}
\label{equa:vectorev}
\end{equation}
where $K$ is the number of layers.
After that, we also follow \citep{xu2019powerful} to sum all the final embeddings of nodes in $\mathcal{G}$ to get the final embedding $\boldsymbol{\mathsf{e}}_{\mathcal{G}}$ of the entire graph $\mathcal{G}$.
We feed $\boldsymbol{\mathsf{e}}_{\mathcal{G}}$ to a single fully-connected layer followed by a $\mathsf{softmax}$ layer to predict the graph label as:
\begin{equation}
\boldsymbol{\mathsf{\hat{y}}}_{\mathcal{G}} = \mathsf{softmax}\left(\textbf{W}\boldsymbol{\mathsf{e}}_{\mathcal{G}} + \textbf{b}\right)
\label{equa:supervisedUGformerloss}
\end{equation}
Finally, we learn the model parameters by minimizing the cross-entropy loss function. 

\subsubsection{Experimental setup}


We use seven well-known datasets consisting of three social network datasets (COLLAB, IMDB-B, and IMDB-M)
and four bioinformatics datasets (DD, MUTAG, PROTEINS, and PTC).
The social network datasets do not have node features; thus, we follow \citep{niepert2016learning,zhang2018end} to use node degrees as features. 
We vary the number $K$ of UGformer layers in \{1, 2, 3\}, the number of steps $T$ in \{1, 2, 3, 4\}, the number of neighbors ($|\mathcal{N}_\mathsf{v}|=N$) sampled for each node in \{4, 8, 16\}, and the hidden size in $\textsc{Trans}(.)$ in \{128, 256, 512, 1024\}.
We set the batch size to 4.
We apply the Adam optimizer \citep{kingma2014adam} to train our UGformer and select the Adam initial learning rate $lr \in \left\{5e^{-5}, 1e^{-4}, 5e^{-4}, 1e^{-3}\right\}$.
We run up to 50 epochs to evaluate our UGformer. 
We follow \citep{xu2019powerful,xinyi2019capsule,maron2019provably,seo2019discriminative,Chen2019ArePG} to use the same data splits and the same 10-fold cross-validation scheme to calculate the classification performance for a fair comparison.
We compare our UGformer with up-to-date strong baselines.
We report the baseline results taken from the original papers or published in \citep{verma2018graph,xinyi2019capsule,FAN2020107084,Chen2019ArePG,seo2019discriminative,xu2019powerful}.

\begin{table}[!ht]
\centering
\caption{Graph classification results (\% accuracy) w.r.t UGformer Variant 1 in an inductive setting.
}
\resizebox{8.5cm}{!}{
\begin{tabular}{l|c|c|c|c|c|c|c}
\hline
{\bf Model} & \textbf{COLLAB} & \textbf{IMDB-B} & \textbf{IMDB-M} & \textbf{DD} & \textbf{PROTEINS} & \textbf{MUTAG} & \textbf{PTC}\\
\hline
PSCN \citep{niepert2016learning} & 72.60 $\pm$ 2.15 & 71.00 $\pm$ 2.29 & 45.23 $\pm$ 2.84 & 77.12 $\pm$ 2.41 & 75.89 $\pm$ 2.76 & \textbf{92.63 $\pm$ 4.21} & 62.29 $\pm$ 5.68\\
GCN \citep{kipf2017semi} & 79.00 $\pm$ 1.80 & 74.00 $\pm$ 3.40 & 51.90 $\pm$ 3.80 & -- & 76.00 $\pm$ 3.20 & 85.60 $\pm$ 5.80 & 64.20 $\pm$ 4.30\\
GFN \citep{Chen2019ArePG} & \textbf{81.50 $\pm$ 2.42} & 73.00 $\pm$ 4.35 & 51.80 $\pm$ 5.16 & 78.78 $\pm$ 3.49 & 76.46 $\pm$ 4.06 & {90.84 $\pm$ 7.22} & -- \\
GraphSAGE \citep{hamilton2017inductive} & 79.70 $\pm$ 1.70 & 72.40 $\pm$ 3.60 & 49.90 $\pm$ 5.00 & 65.80 $\pm$ 4.90 & 65.90 $\pm$ 2.70 & 79.80 $\pm$ 13.9 & -- \\
GAT \citep{velickovic2018graph} & 75.80 $\pm$ 1.60 & 70.50 $\pm$ 2.30 & 47.80 $\pm$ 3.10 & -- & 74.70 $\pm$ 2.20 & 89.40 $\pm$ 6.10 & 66.70 $\pm$ 5.10 \\
DGCNN \citep{zhang2018end} & 73.76 $\pm$ 0.49 & 70.03 $\pm$ 0.86 & 47.83 $\pm$ 0.85 & 79.37 $\pm$ 0.94 & 75.54 $\pm$ 0.94 & 85.83 $\pm$ 1.66 & 58.59 $\pm$ 2.47\\
SAGPool \citep{Lee2019SelfAttentionGP} & -- & -- & -- & 76.45 $\pm$ 0.97 & 71.86 $\pm$ 0.97 & -- & -- \\
PPGN \citep{maron2019provably} & 81.38 $\pm$ 1.42 & 73.00 $\pm$ 5.77 & 50.46 $\pm$ 3.59 & -- & 77.20 $\pm$ 4.73 & 90.55 $\pm$ 8.70 & 66.17 $\pm$ 6.54\\
CapsGNN \citep{xinyi2019capsule} & 79.62 $\pm$ 0.91 & 73.10 $\pm$ 4.83 & 50.27 $\pm$ 2.65 & 75.38 $\pm$ 4.17 & 76.28 $\pm$ 3.63 & 86.67 $\pm$ 6.88 & -- \\
DSGC \citep{seo2019discriminative} & 79.20 $\pm$ 1.60 & 73.20 $\pm$ 4.90 & 48.50 $\pm$ 4.80 & 77.40 $\pm$ 6.40 & 74.20 $\pm$ 3.80 & 86.70 $\pm$ 7.60 & --\\
GCAPS \citep{verma2018graph} & 77.71 $\pm$ 2.51 & 71.69 $\pm$ 3.40 & 48.50 $\pm$ 4.10 & 77.62 $\pm$ 4.99 & 76.40 $\pm$ 4.17 & -- & 66.01 $\pm$ 5.91\\
IEGN \citep{maron2019invariant} & 77.92 $\pm$ 1.70 & 71.27 $\pm$ 4.50 & 48.55 $\pm$ 3.90 & -- & 75.19 $\pm$ 4.30 & 84.61 $\pm$ 10.0 & 59.47 $\pm$ 7.30 \\
GIN-0 \citep{xu2019powerful} & 80.20 $\pm$ 1.90 & 75.10 $\pm$ 5.10 & 52.30 $\pm$ 2.80 & -- & 76.20 $\pm$ 2.80 & {89.40 $\pm$ 5.60} & 64.60 $\pm$ 7.00\\
\hline
\textbf{UGformer} & 77.84 $\pm$ 1.48 & \textbf{77.04 $\pm$ 3.45} & \textbf{53.60 $\pm$ 3.53} & \textbf{80.23 $\pm$ 1.48} & \textbf{78.53 $\pm$ 4.07} & 89.97 $\pm$ 3.65 & \textbf{69.63 $\pm$ 3.60}\\
\hline
\end{tabular}
}
\label{tab:expresult_sup}
\end{table}

\subsubsection{Main results} 
\label{subsec:expresults}
Table \ref{tab:expresult_sup} presents the experimental results of UGformer and other strong baseline models for the benchmark datasets.
In general, our UGformer gains competitive accuracies on the social network datasets. 
Especially, UGformer produces state-of-the-art accuracies 
on IMDB-B and IMDB-M respectively, which outperform those of other existing models. 
On the bioinformatics datasets, UGformer obtains the highest accuracies 
on DD, PROTEINS, and PTC, respectively.
Besides, there are no significant differences between our UGformer and the baselines on MUTAG as this dataset only consists of 188 graphs.

\subsection{UGformer Variant 1 for graph classification in an ``unsupervised transductive'' setting}

We propose an \textit{unsupervised transductive} learning approach to train GNNs to address the limited availability of class labels.
We consider a final embedding $\boldsymbol{\mathsf{o}}_{\mathsf{v}}$ for each node $\mathsf{v}$, and make the similarity between $\boldsymbol{\mathsf{e}}_\mathsf{v}$ and $\boldsymbol{\mathsf{o}}_{\mathsf{v}}$ higher than that between $\boldsymbol{\mathsf{e}}_\mathsf{v}$ and the final embeddings of the other nodes. 
The goal is to guide GNNs to recognize and distinguish the sub-graph structural information within each graph and also memorize the structural differences among graphs to produce better node and graph embeddings.
We aim to minimize the sampled $\mathsf{softmax}$ loss function \citep{Jean2015} applied to node $\mathsf{v}$ as:
\begin{equation}
\mathcal{L}_{\mathsf{GNN}}\left(\mathsf{v}\right) = -\log \frac{\exp(\boldsymbol{\mathsf{o}}_\mathsf{v}^\mathsf{T}\boldsymbol{\mathsf{e}}_{\mathsf{v}})}{\sum_{\mathsf{v'} \in \mathcal{V'}} \exp(\boldsymbol{\mathsf{o}}_\mathsf{v'}^\mathsf{T}\boldsymbol{\mathsf{e}}_{\mathsf{v}})}
\end{equation}
where $\mathcal{V'}$ is a subset sampled from $\left\{\cup\mathcal{V}_m\right\}_{m=1}^M$. 
Node embeddings $\boldsymbol{\mathsf{o}}_{\mathsf{v}}$ are also learned as model parameters. We then sum all the final embeddings $\boldsymbol{\mathsf{o}}_{\mathsf{v}}$ of nodes $\mathsf{v}$ in $\mathcal{G}$ to obtain the graph embedding $\boldsymbol{\mathsf{e}}_{\mathcal{G}}$. 

\subsubsection{Experimental setup} 
We use the same seven well-known datasets as used in the inductive setting.
We 
follow some unsupervised approaches such as DGK \citep {yanardag2015deep} and AWE \citep{ivanov2018anonymous} to train our unsupervised UGformer on all nodes from the entire dataset (i.e., {consisting of all nodes from the test set during training}).
The hyper-parameters are also varied as same as in the inductive setting.
We also train our 
GCN baseline following our unsupervised learning.
We set the batch size to 4 and vary the number of GCN layers in \{1, 2, 3\} and the hidden layer size in \{32, 64, 128, 256\}. We also use the Adam optimizer \citep{kingma2014adam} to train this unsupervised GCN up to 50 epochs.

We finally utilize the logistic regression classifier \citep{Fan:2008} with using the 10-fold cross-validation scheme to evaluate our models.
We compare our unsupervised GCN (denoted as uGCN) and UGformer with the baselines: Deep Graph Kernel (DGK) \citep{yanardag2015deep} and Anonymous Walk Embedding (AWE) \citep{ivanov2018anonymous}.

\begin{table}[!ht]
\caption{Graph classification results (\% accuracy) w.r.t UGformer Variant 1 in an unsupervised transductive setting. 
Note that we do not aim to directly compare the unsupervised transductive setting with the inductive setting.}
\centering
\resizebox{8.5cm}{!}{
\begin{tabular}{l|c|c|c|c|c|c|c}
\hline
\bf Model &  \textbf{COLLAB} &  \textbf{IMDB-B} &  \textbf{IMDB-M} & \textbf{DD} & \textbf{PROTEINS} & \textbf{MUTAG} & \textbf{PTC}\\
\hline
DGK \citeyearpar{yanardag2015deep} & 73.09 $\pm$ 0.25 & 66.96 $\pm$ 0.56 & 44.55 $\pm$ 0.52 & 73.50 $\pm$ 1.01 & 75.68 $\pm$ 0.54 & 87.44 $\pm$ 2.72 & 60.08 $\pm$ 2.55\\
AWE \citeyearpar{ivanov2018anonymous} & 73.93 $\pm$ 1.94 & 74.45 $\pm$ 5.83 & 51.54 $\pm$ 3.61 & 71.51 $\pm$ 4.02 & -- & {87.87 $\pm$ 9.76} & --\\
\hline
uGCN & 93.28 $\pm$ 0.99 & 94.50 $\pm$ 2.79 & {81.66 $\pm$ 3.16} & 94.31 $\pm$ 1.71 & \textbf{89.09 $\pm$ 3.25} & \textbf{95.36 $\pm$ 2.64} & \textbf{92.67 $\pm$ 4.60}\\
\textbf{UGformer} & \textbf{95.62 $\pm$ 0.92} & \textbf{96.41 $\pm$ 1.94} & \textbf{89.20 $\pm$ 2.52} & \textbf{95.67 $\pm$ 1.89} & {80.01 $\pm$ 3.21} & 88.47 $\pm$ 7.13 & {91.81 $\pm$ 6.61}\\
\hline
\end{tabular}
}
\label{tab:expresult_unsup}
\end{table}

\subsubsection{Main results} 
Table \ref{tab:expresult_unsup} presents the experimental results in the unsupervised transductive setting, wherein both our unsupervised UGformer and uGCN obtain new state-of-the-art accuracies on the benchmark datasets.
The significant gains demonstrate a notable impact of our unsupervised transductive learning approach.

\section{Conclusion}
\label{sec:conclusion}

We present a transformer-based GNN model, named UGformer, to learn graph representations.
We consider two UGformer variants of (i) leveraging the transformer on a set of sampled neighbors for each input node and (ii) leveraging the transformer on all input nodes. 
Experimental results show that our graph transformer UGformer produces state-of-the-art accuracies on well-known benchmark datasets for graph classification and text classification.
Furthermore, we hope that future GNN works can consider the unsupervised transductive setting to address the limited availability of class labels.


\bibliographystyle{ACM-Reference-Format}
\bibliography{references}


\begin{thebibliography}{34}


\ifx \showCODEN    \undefined \def \showCODEN     #1{\unskip}     \fi
\ifx \showDOI      \undefined \def \showDOI       #1{#1}\fi
\ifx \showISBNx    \undefined \def \showISBNx     #1{\unskip}     \fi
\ifx \showISBNxiii \undefined \def \showISBNxiii  #1{\unskip}     \fi
\ifx \showISSN     \undefined \def \showISSN      #1{\unskip}     \fi
\ifx \showLCCN     \undefined \def \showLCCN      #1{\unskip}     \fi
\ifx \shownote     \undefined \def \shownote      #1{#1}          \fi
\ifx \showarticletitle \undefined \def \showarticletitle #1{#1}   \fi
\ifx \showURL      \undefined \def \showURL       {\relax}        \fi
\providecommand\bibfield[2]{#2}
\providecommand\bibinfo[2]{#2}
\providecommand\natexlab[1]{#1}
\providecommand\showeprint[2][]{arXiv:#2}

\bibitem[Chen et~al\mbox{.}(2019)]%
        {Chen2019ArePG}
\bibfield{author}{\bibinfo{person}{Ting Chen}, \bibinfo{person}{Song Bian},
  {and} \bibinfo{person}{Yizhou Sun}.} \bibinfo{year}{2019}\natexlab{}.
\newblock \showarticletitle{Are Powerful Graph Neural Nets Necessary? A
  Dissection on Graph Classification}.
\newblock \bibinfo{journal}{\emph{arXiv:1905.04579}} (\bibinfo{year}{2019}).
\newblock


\bibitem[Fan et~al\mbox{.}(2008)]%
        {Fan:2008}
\bibfield{author}{\bibinfo{person}{Rong-En Fan}, \bibinfo{person}{Kai-Wei
  Chang}, \bibinfo{person}{Cho-Jui Hsieh}, \bibinfo{person}{Xiang-Rui Wang},
  {and} \bibinfo{person}{Chih-Jen Lin}.} \bibinfo{year}{2008}\natexlab{}.
\newblock \showarticletitle{{LIBLINEAR: A Library for Large Linear
  Classification}}.
\newblock \bibinfo{journal}{\emph{Journal of Machine Learning Research}}
  \bibinfo{volume}{9} (\bibinfo{year}{2008}), \bibinfo{pages}{1871--1874}.
\newblock


\bibitem[Fan et~al\mbox{.}(2019)]%
        {FAN2020107084}
\bibfield{author}{\bibinfo{person}{Xiaolong Fan}, \bibinfo{person}{Maoguo
  Gong}, \bibinfo{person}{Yu Xie}, \bibinfo{person}{Fenlong Jiang}, {and}
  \bibinfo{person}{Hao Li}.} \bibinfo{year}{2019}\natexlab{}.
\newblock \showarticletitle{Structured Self-attention Architecture for
  Graph-level Representation Learning}.
\newblock \bibinfo{journal}{\emph{Pattern Recognition}}  \bibinfo{volume}{100}
  (\bibinfo{year}{2019}).
\newblock
\showISSN{0031-3203}


\bibitem[Gilmer et~al\mbox{.}(2017)]%
        {Gilmer2017}
\bibfield{author}{\bibinfo{person}{Justin Gilmer}, \bibinfo{person}{Samuel~S.
  Schoenholz}, \bibinfo{person}{Patrick~F. Riley}, \bibinfo{person}{Oriol
  Vinyals}, {and} \bibinfo{person}{George~E. Dahl}.}
  \bibinfo{year}{2017}\natexlab{}.
\newblock \showarticletitle{{Neural Message Passing for Quantum Chemistry}}. In
  \bibinfo{booktitle}{\emph{ICML}}.
\newblock


\bibitem[Hamilton et~al\mbox{.}(2017a)]%
        {hamilton2017inductive}
\bibfield{author}{\bibinfo{person}{William~L. Hamilton}, \bibinfo{person}{Rex
  Ying}, {and} \bibinfo{person}{Jure Leskovec}.}
  \bibinfo{year}{2017}\natexlab{a}.
\newblock \showarticletitle{Inductive representation learning on large graphs}.
  In \bibinfo{booktitle}{\emph{NeurIPS}}. \bibinfo{pages}{1024--1034}.
\newblock


\bibitem[Hamilton et~al\mbox{.}(2017b)]%
        {hamilton2017representation}
\bibfield{author}{\bibinfo{person}{William~L. Hamilton}, \bibinfo{person}{Rex
  Ying}, {and} \bibinfo{person}{Jure Leskovec}.}
  \bibinfo{year}{2017}\natexlab{b}.
\newblock \showarticletitle{Representation learning on graphs: Methods and
  applications}.
\newblock \bibinfo{journal}{\emph{arXiv:1709.05584}} (\bibinfo{year}{2017}).
\newblock


\bibitem[Ivanov and Burnaev(2018)]%
        {ivanov2018anonymous}
\bibfield{author}{\bibinfo{person}{Sergey Ivanov} {and} \bibinfo{person}{Evgeny
  Burnaev}.} \bibinfo{year}{2018}\natexlab{}.
\newblock \showarticletitle{Anonymous Walk Embeddings}. In
  \bibinfo{booktitle}{\emph{ICML}}. \bibinfo{pages}{2191--2200}.
\newblock


\bibitem[Jean et~al\mbox{.}(2015)]%
        {Jean2015}
\bibfield{author}{\bibinfo{person}{S{\'e}bastien Jean},
  \bibinfo{person}{Kyunghyun Cho}, \bibinfo{person}{Roland Memisevic}, {and}
  \bibinfo{person}{Yoshua Bengio}.} \bibinfo{year}{2015}\natexlab{}.
\newblock \showarticletitle{On Using Very Large Target Vocabulary for Neural
  Machine Translation}. In \bibinfo{booktitle}{\emph{ACL}}.
\newblock


\bibitem[Kingma and Ba(2015)]%
        {kingma2014adam}
\bibfield{author}{\bibinfo{person}{Diederik Kingma} {and}
  \bibinfo{person}{Jimmy Ba}.} \bibinfo{year}{2015}\natexlab{}.
\newblock \showarticletitle{Adam: A method for stochastic optimization}.
\newblock \bibinfo{journal}{\emph{ICLR}} (\bibinfo{year}{2015}).
\newblock


\bibitem[Kipf and Welling(2017)]%
        {kipf2017semi}
\bibfield{author}{\bibinfo{person}{Thomas~N. Kipf} {and} \bibinfo{person}{Max
  Welling}.} \bibinfo{year}{2017}\natexlab{}.
\newblock \showarticletitle{Semi-Supervised Classification with Graph
  Convolutional Networks}. In \bibinfo{booktitle}{\emph{ICLR}}.
\newblock


\bibitem[Lee et~al\mbox{.}(2019)]%
        {Lee2019SelfAttentionGP}
\bibfield{author}{\bibinfo{person}{Junhyun Lee}, \bibinfo{person}{Inyeop Lee},
  {and} \bibinfo{person}{Jaewoo Kang}.} \bibinfo{year}{2019}\natexlab{}.
\newblock \showarticletitle{Self-Attention Graph Pooling}. In
  \bibinfo{booktitle}{\emph{ICML}}. \bibinfo{pages}{3734--3743}.
\newblock


\bibitem[Li et~al\mbox{.}(2016)]%
        {li2015gated}
\bibfield{author}{\bibinfo{person}{Yujia Li}, \bibinfo{person}{Daniel Tarlow},
  \bibinfo{person}{Marc Brockschmidt}, {and} \bibinfo{person}{Richard Zemel}.}
  \bibinfo{year}{2016}\natexlab{}.
\newblock \showarticletitle{{Gated Graph Sequence Neural Networks}}.
\newblock \bibinfo{journal}{\emph{ICLR}} (\bibinfo{year}{2016}).
\newblock


\bibitem[Maron et~al\mbox{.}(2019a)]%
        {maron2019provably}
\bibfield{author}{\bibinfo{person}{Haggai Maron}, \bibinfo{person}{Heli
  Ben-Hamu}, \bibinfo{person}{Hadar Serviansky}, {and} \bibinfo{person}{Yaron
  Lipman}.} \bibinfo{year}{2019}\natexlab{a}.
\newblock \showarticletitle{Provably Powerful Graph Networks}. In
  \bibinfo{booktitle}{\emph{NeurIPS}}. \bibinfo{pages}{2153--2164}.
\newblock


\bibitem[Maron et~al\mbox{.}(2019b)]%
        {maron2019invariant}
\bibfield{author}{\bibinfo{person}{Haggai Maron}, \bibinfo{person}{Heli
  Ben-Hamu}, \bibinfo{person}{Nadav Shamir}, {and} \bibinfo{person}{Yaron
  Lipman}.} \bibinfo{year}{2019}\natexlab{b}.
\newblock \showarticletitle{Invariant and equivariant graph networks}.
\newblock \bibinfo{journal}{\emph{ICLR}} (\bibinfo{year}{2019}).
\newblock


\bibitem[Nguyen(2021)]%
        {NGUYEN2021Thesis}
\bibfield{author}{\bibinfo{person}{Dai~Quoc Nguyen}.}
  \bibinfo{year}{2021}\natexlab{}.
\newblock \emph{\bibinfo{title}{{Representation Learning for Graph-Structured
  Data}}}.
\newblock \bibinfo{thesistype}{Ph.\,D. Dissertation}. \bibinfo{school}{Monash
  University}.
\newblock
\urldef\tempurl%
\url{https://doi.org/10.26180/14450496.v1}
\showDOI{\tempurl}


\bibitem[Nguyen et~al\mbox{.}(2021)]%
        {Nguyen2020QGNN}
\bibfield{author}{\bibinfo{person}{Dai~Quoc Nguyen}, \bibinfo{person}{Tu~Dinh
  Nguyen}, {and} \bibinfo{person}{Dinh Phung}.}
  \bibinfo{year}{2021}\natexlab{}.
\newblock \showarticletitle{Quaternion Graph Neural Networks}. In
  \bibinfo{booktitle}{\emph{Asian Conference on Machine Learning}}.
\newblock


\bibitem[Nguyen et~al\mbox{.}(2022b)]%
        {Nguyen2022NoGE}
\bibfield{author}{\bibinfo{person}{Dai~Quoc Nguyen}, \bibinfo{person}{Vinh
  Tong}, \bibinfo{person}{Dinh Phung}, {and} \bibinfo{person}{Dat~Quoc
  Nguyen}.} \bibinfo{year}{2022}\natexlab{b}.
\newblock \showarticletitle{Node Co-occurrence based Graph Neural Networks for
  Knowledge Graph Link Prediction}. In \bibinfo{booktitle}{\emph{Proceedings of
  WSDM 2022 (Demonstrations)}}.
\newblock


\bibitem[Nguyen et~al\mbox{.}(2022a)]%
        {Nguyen2021regvd}
\bibfield{author}{\bibinfo{person}{Van-Anh Nguyen}, \bibinfo{person}{Dai~Quoc
  Nguyen}, \bibinfo{person}{Van Nguyen}, \bibinfo{person}{Trung Le},
  \bibinfo{person}{Quan~Hung Tran}, {and} \bibinfo{person}{Dinh Phung}.}
  \bibinfo{year}{2022}\natexlab{a}.
\newblock \showarticletitle{ReGVD: Revisiting Graph Neural Networks for
  Vulnerability Detection}. In \bibinfo{booktitle}{\emph{Proceedings of ICSE
  2022 (Demonstrations)}}.
\newblock


\bibitem[Niepert et~al\mbox{.}(2016)]%
        {niepert2016learning}
\bibfield{author}{\bibinfo{person}{Mathias Niepert}, \bibinfo{person}{Mohamed
  Ahmed}, {and} \bibinfo{person}{Konstantin Kutzkov}.}
  \bibinfo{year}{2016}\natexlab{}.
\newblock \showarticletitle{{Learning Convolutional Neural Networks for
  Graphs}}. In \bibinfo{booktitle}{\emph{ICML}}. \bibinfo{pages}{2014--2023}.
\newblock


\bibitem[Pennington et~al\mbox{.}(2014)]%
        {pennington2014glove}
\bibfield{author}{\bibinfo{person}{Jeffrey Pennington},
  \bibinfo{person}{Richard Socher}, {and} \bibinfo{person}{Christopher~D.
  Manning}.} \bibinfo{year}{2014}\natexlab{}.
\newblock \showarticletitle{{GloVe: Global Vectors for Word Representation}}.
  In \bibinfo{booktitle}{\emph{Proceedings of the 2014 Conference on Empirical
  Methods in Natural Language Processing}}. \bibinfo{pages}{1532--1543}.
\newblock


\bibitem[Scarselli et~al\mbox{.}(2009)]%
        {scarselli2009graph}
\bibfield{author}{\bibinfo{person}{Franco Scarselli}, \bibinfo{person}{Marco
  Gori}, \bibinfo{person}{Ah~Chung Tsoi}, \bibinfo{person}{Markus
  Hagenbuchner}, {and} \bibinfo{person}{Gabriele Monfardini}.}
  \bibinfo{year}{2009}\natexlab{}.
\newblock \showarticletitle{The graph neural network model}.
\newblock \bibinfo{journal}{\emph{IEEE Transactions on Neural Networks}}
  \bibinfo{volume}{20}, \bibinfo{number}{1} (\bibinfo{year}{2009}),
  \bibinfo{pages}{61--80}.
\newblock


\bibitem[Seo et~al\mbox{.}(2019)]%
        {seo2019discriminative}
\bibfield{author}{\bibinfo{person}{Younjoo Seo}, \bibinfo{person}{Andreas
  Loukas}, {and} \bibinfo{person}{Nathanael Peraudin}.}
  \bibinfo{year}{2019}\natexlab{}.
\newblock \showarticletitle{Discriminative structural graph classification}.
\newblock \bibinfo{journal}{\emph{arXiv:1905.13422}} (\bibinfo{year}{2019}).
\newblock


\bibitem[Vaswani et~al\mbox{.}(2017)]%
        {vaswani2017attention}
\bibfield{author}{\bibinfo{person}{Ashish Vaswani}, \bibinfo{person}{Noam
  Shazeer}, \bibinfo{person}{Niki Parmar}, \bibinfo{person}{Jakob Uszkoreit},
  \bibinfo{person}{Llion Jones}, \bibinfo{person}{Aidan~N Gomez},
  \bibinfo{person}{{\L}ukasz Kaiser}, {and} \bibinfo{person}{Illia
  Polosukhin}.} \bibinfo{year}{2017}\natexlab{}.
\newblock \showarticletitle{Attention is all you need}. In
  \bibinfo{booktitle}{\emph{NeurIPS}}. \bibinfo{pages}{5998--6008}.
\newblock


\bibitem[Veli{\v{c}}kovi{\'{c}} et~al\mbox{.}(2018)]%
        {velickovic2018graph}
\bibfield{author}{\bibinfo{person}{Petar Veli{\v{c}}kovi{\'{c}}},
  \bibinfo{person}{Guillem Cucurull}, \bibinfo{person}{Arantxa Casanova},
  \bibinfo{person}{Adriana Romero}, \bibinfo{person}{Pietro Li{\`{o}}}, {and}
  \bibinfo{person}{Yoshua Bengio}.} \bibinfo{year}{2018}\natexlab{}.
\newblock \showarticletitle{{Graph Attention Networks}}.
\newblock \bibinfo{journal}{\emph{ICLR}} (\bibinfo{year}{2018}).
\newblock


\bibitem[Verma and Zhang(2018)]%
        {verma2018graph}
\bibfield{author}{\bibinfo{person}{Saurabh Verma} {and} \bibinfo{person}{Zhi-Li
  Zhang}.} \bibinfo{year}{2018}\natexlab{}.
\newblock \showarticletitle{Graph capsule convolutional neural networks}.
\newblock \bibinfo{journal}{\emph{The Joint ICML and IJCAI Workshop on
  Computational Biology}} (\bibinfo{year}{2018}).
\newblock


\bibitem[Wu et~al\mbox{.}(2019)]%
        {wu2019comprehensive}
\bibfield{author}{\bibinfo{person}{Zonghan Wu}, \bibinfo{person}{Shirui Pan},
  \bibinfo{person}{Fengwen Chen}, \bibinfo{person}{Guodong Long},
  \bibinfo{person}{Chengqi Zhang}, {and} \bibinfo{person}{Philip~S Yu}.}
  \bibinfo{year}{2019}\natexlab{}.
\newblock \showarticletitle{A comprehensive survey on graph neural networks}.
\newblock \bibinfo{journal}{\emph{arXiv:1901.00596}} (\bibinfo{year}{2019}).
\newblock


\bibitem[Xinyi and Chen(2019)]%
        {xinyi2019capsule}
\bibfield{author}{\bibinfo{person}{Zhang Xinyi} {and} \bibinfo{person}{Lihui
  Chen}.} \bibinfo{year}{2019}\natexlab{}.
\newblock \showarticletitle{{Capsule Graph Neural Network}}.
\newblock \bibinfo{journal}{\emph{ICLR}} (\bibinfo{year}{2019}).
\newblock


\bibitem[Xu et~al\mbox{.}(2019)]%
        {xu2019powerful}
\bibfield{author}{\bibinfo{person}{Keyulu Xu}, \bibinfo{person}{Weihua Hu},
  \bibinfo{person}{Jure Leskovec}, {and} \bibinfo{person}{Stefanie Jegelka}.}
  \bibinfo{year}{2019}\natexlab{}.
\newblock \showarticletitle{{How Powerful Are Graph Neural Networks?}}
\newblock \bibinfo{journal}{\emph{ICLR}} (\bibinfo{year}{2019}).
\newblock


\bibitem[Xu et~al\mbox{.}(2018)]%
        {xu2018representation}
\bibfield{author}{\bibinfo{person}{Keyulu Xu}, \bibinfo{person}{Chengtao Li},
  \bibinfo{person}{Yonglong Tian}, \bibinfo{person}{Tomohiro Sonobe},
  \bibinfo{person}{Ken-ichi Kawarabayashi}, {and} \bibinfo{person}{Stefanie
  Jegelka}.} \bibinfo{year}{2018}\natexlab{}.
\newblock \showarticletitle{Representation Learning on Graphs with Jumping
  Knowledge Networks}. In \bibinfo{booktitle}{\emph{ICML}}.
  \bibinfo{pages}{5453--5462}.
\newblock


\bibitem[Yanardag and Vishwanathan(2015)]%
        {yanardag2015deep}
\bibfield{author}{\bibinfo{person}{Pinar Yanardag} {and} \bibinfo{person}{SVN
  Vishwanathan}.} \bibinfo{year}{2015}\natexlab{}.
\newblock \showarticletitle{Deep Graph Kernels}. In
  \bibinfo{booktitle}{\emph{The ACM SIGKDD Conference on Knowledge Discovery
  and Data Mining}}. \bibinfo{pages}{1365--1374}.
\newblock


\bibitem[Yao et~al\mbox{.}(2019)]%
        {yao2019graph}
\bibfield{author}{\bibinfo{person}{Liang Yao}, \bibinfo{person}{Chengsheng
  Mao}, {and} \bibinfo{person}{Yuan Luo}.} \bibinfo{year}{2019}\natexlab{}.
\newblock \showarticletitle{Graph convolutional networks for text
  classification}. In \bibinfo{booktitle}{\emph{AAAI}},
  Vol.~\bibinfo{volume}{33}. \bibinfo{pages}{7370--7377}.
\newblock


\bibitem[Ying et~al\mbox{.}(2018)]%
        {Ying2018diffpool}
\bibfield{author}{\bibinfo{person}{Rex Ying}, \bibinfo{person}{Jiaxuan You},
  \bibinfo{person}{Christopher Morris}, \bibinfo{person}{Xiang Ren},
  \bibinfo{person}{William~L. Hamilton}, {and} \bibinfo{person}{Jure
  Leskovec}.} \bibinfo{year}{2018}\natexlab{}.
\newblock \showarticletitle{Hierarchical Graph Representation Learning with
  Differentiable Pooling}. In \bibinfo{booktitle}{\emph{NeurIPS}}.
  \bibinfo{pages}{4805--4815}.
\newblock


\bibitem[Zhang et~al\mbox{.}(2018)]%
        {zhang2018end}
\bibfield{author}{\bibinfo{person}{Muhan Zhang}, \bibinfo{person}{Zhicheng
  Cui}, \bibinfo{person}{Marion Neumann}, {and} \bibinfo{person}{Yixin Chen}.}
  \bibinfo{year}{2018}\natexlab{}.
\newblock \showarticletitle{{An End-to-End Deep Learning Architecture for Graph
  Classification}}. In \bibinfo{booktitle}{\emph{AAAI}}.
\newblock


\bibitem[Zhang et~al\mbox{.}(2020)]%
        {zhang2020every}
\bibfield{author}{\bibinfo{person}{Yufeng Zhang}, \bibinfo{person}{Xueli Yu},
  \bibinfo{person}{Zeyu Cui}, \bibinfo{person}{Shu Wu},
  \bibinfo{person}{Zhongzhen Wen}, {and} \bibinfo{person}{Liang Wang}.}
  \bibinfo{year}{2020}\natexlab{}.
\newblock \showarticletitle{Every Document Owns Its Structure: Inductive Text
  Classification via Graph Neural Networks}. In
  \bibinfo{booktitle}{\emph{ACL}}. \bibinfo{pages}{334--339}.
\newblock


\end{thebibliography}

\end{document}